\documentclass[sigconf]{acmart}

 \AtBeginDocument{
 \settopmatter{printacmref=false} 
 \renewcommand\footnotetextcopyrightpermission[1]{} 
 \pagestyle{plain}\pagenumbering{gobble} 
}

\usepackage[draft=false]{defs-common}
\usepackage{defs-custom}

\usepackage{algorithm}
\usepackage{algorithmic}
\usepackage{multirow}

\copyrightyear{2021}
\acmYear{2021}
\setcopyright{acmcopyright}\acmConference[FDG'21]{The 16th International Conference on the Foundations of Digital Games (FDG) 2021}{August 3--6, 2021}{Montreal, QC, Canada}
\acmBooktitle{The 16th International Conference on the Foundations of Digital Games (FDG) 2021 (FDG'21), August 3--6, 2021, Montreal, QC, Canada}
\acmPrice{15.00}
\acmDOI{10.1145/3472538.3472545}
\acmISBN{978-1-4503-8422-3/21/08}

\begin{document}

\title{Generating and Blending Game Levels via Quality-Diversity in the Latent Space of a Variational Autoencoder}

\author{Anurag Sarkar}
\affiliation{
  \institution{Northeastern University}
  \city{Boston}
  \state{MA}
  \country{USA}}
\email{sarkar.an@northeastern.edu}

\author{Seth Cooper}
\affiliation{
  \institution{Northeastern University}
  \city{Boston}
  \state{MA}
  \country{USA}}
\email{se.cooper@northeastern.edu}

\begin{abstract}
Several works have demonstrated the use of variational autoencoders (VAEs) for generating levels in the style of existing games and blending levels across different games. Further, quality-diversity (QD) algorithms have also become popular for generating varied game content by using evolution to explore a search space while focusing on both variety and quality. To reap the benefits of both these approaches, we present a level generation and game blending approach that combines the use of VAEs and QD algorithms. Specifically, we train VAEs on game levels and run the MAP-Elites QD algorithm using the learned latent space of the VAE as the search space. The latent space captures the properties of the games whose levels we want to generate and blend, while MAP-Elites searches this latent space to find a diverse set of levels optimizing a given objective such as playability. We test our method using models for 5 different platformer games as well as a blended domain spanning 3 of these games. We refer to using MAP-Elites for blending as  \textit{Blend-Elites}. Our results show that MAP-Elites in conjunction with VAEs enables the generation of a diverse set of playable levels not just for each individual game but also for the blended domain while illuminating game-specific regions of the blended latent space.
\end{abstract}


\maketitle


\newcommand{\XFIGUREdenl}{
\begin{figure*}[t!]
\centering
\begin{subfigure}[t]{0.29\textwidth}
\centering
\includegraphics[width=1\linewidth]{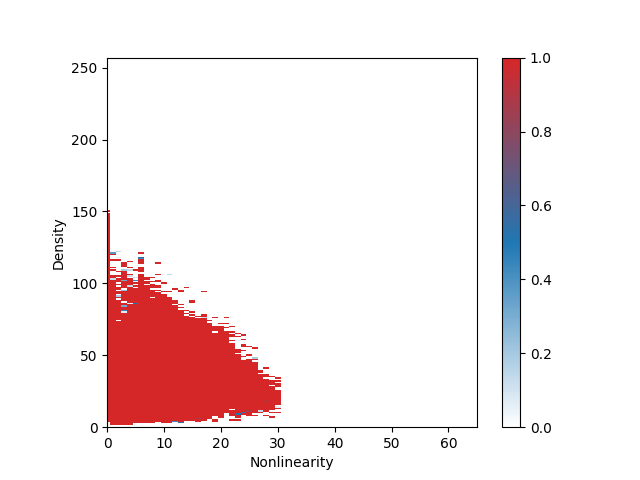}
\caption{SMB}
\end{subfigure}
~
\begin{subfigure}[t]{0.29\textwidth}
\centering
\includegraphics[width=1\linewidth]{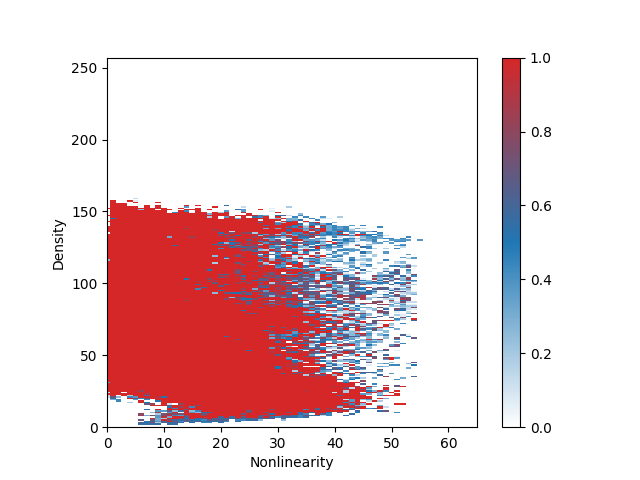}
\caption{KI}
\end{subfigure}
~
\begin{subfigure}[t]{0.29\textwidth}
\includegraphics[width=1\linewidth]{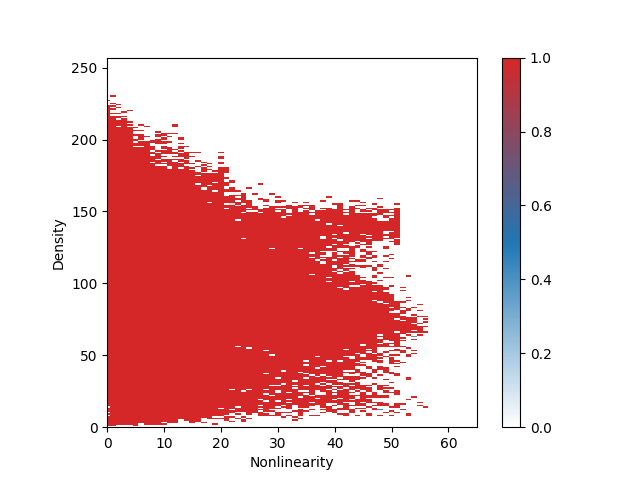}
\caption{MM}
\end{subfigure}
\\
\begin{subfigure}{0.29\textwidth}
\includegraphics[width=1\linewidth]{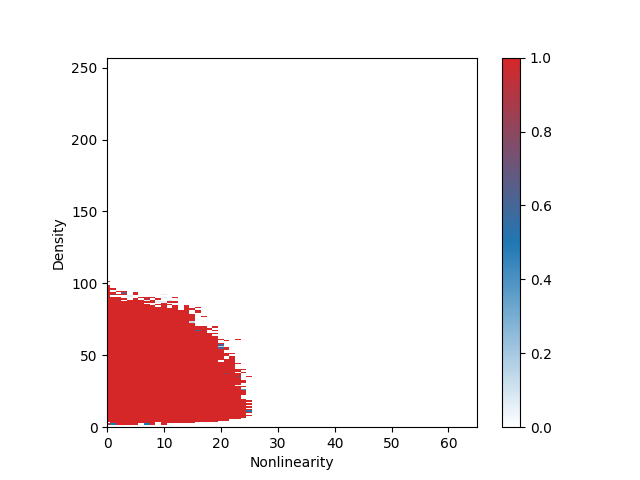}
\caption{CV}
\end{subfigure}
~
\begin{subfigure}{0.29\textwidth}
\includegraphics[width=1\linewidth]{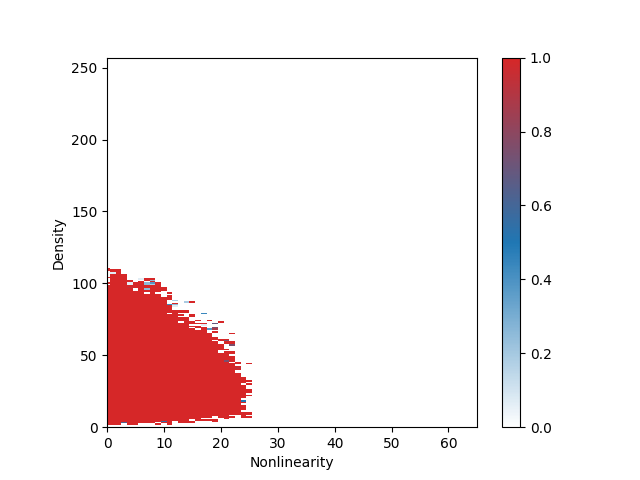}
\caption{NG}
\end{subfigure}
~
\begin{subfigure}{0.29\textwidth}
\includegraphics[width=1\linewidth]{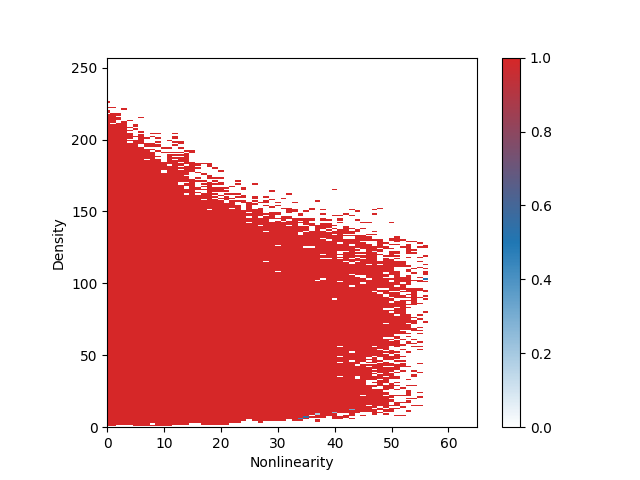}
\caption{SMB-KI-MM}
\end{subfigure}
\caption{\label{XFIGUREdenl} Archives for the Density-Nonlinearity BC.}
\end{figure*}
}

\newcommand{\XFIGUREsymsim}{
\begin{figure*}[t!]
\centering
\begin{subfigure}[t]{0.29\textwidth}
\centering
\includegraphics[width=1\linewidth]{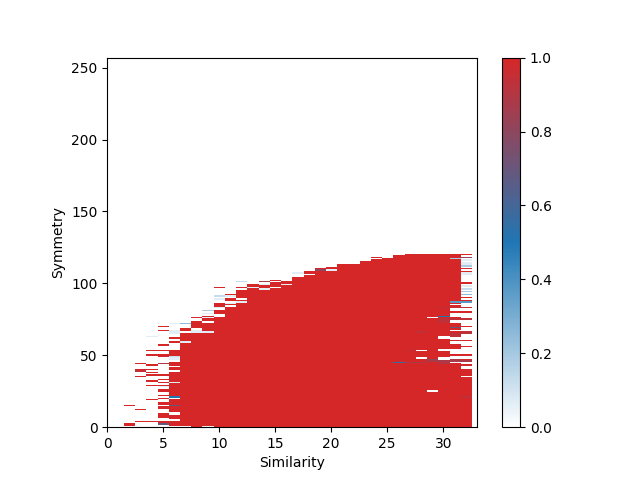}
\caption{SMB}
\end{subfigure}
~
\begin{subfigure}[t]{0.29\textwidth}
\centering
\includegraphics[width=1\linewidth]{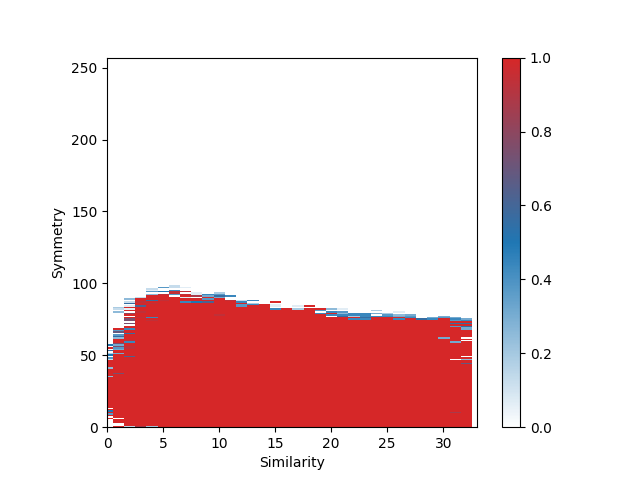}
\caption{KI}
\end{subfigure}
~
\begin{subfigure}[t]{0.29\textwidth}
\includegraphics[width=1\linewidth]{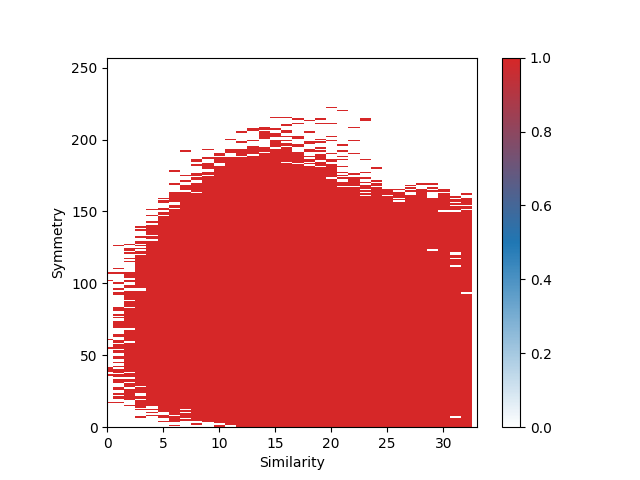}
\caption{MM}
\end{subfigure}
\\
\begin{subfigure}{0.29\textwidth}
\includegraphics[width=1\linewidth]{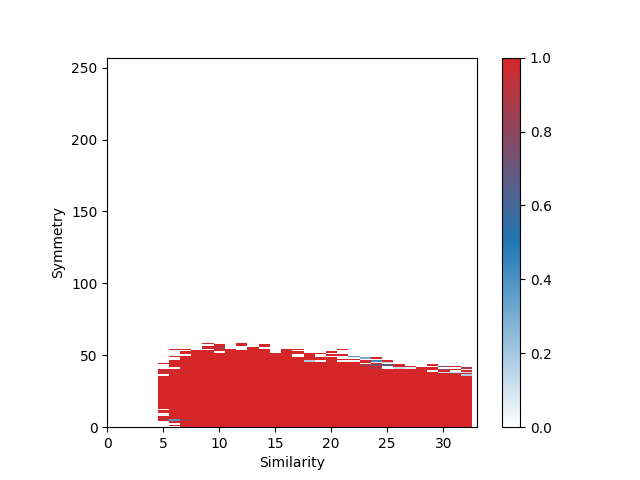}
\caption{CV}
\end{subfigure}
~
\begin{subfigure}{0.29\textwidth}
\includegraphics[width=1\linewidth]{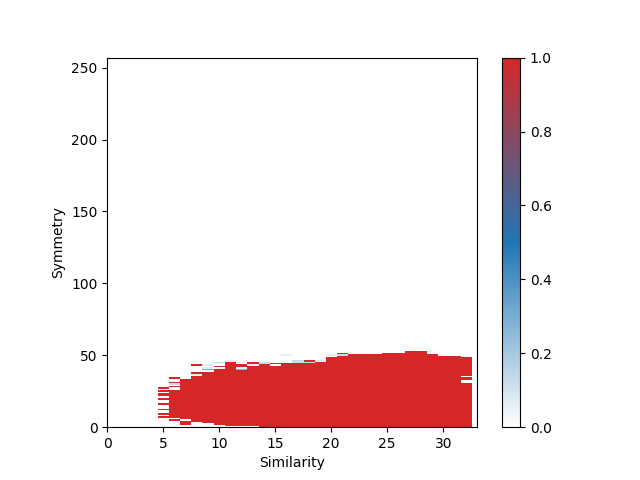}
\caption{NG}
\end{subfigure}
~
\begin{subfigure}{0.29\textwidth}
\includegraphics[width=1\linewidth]{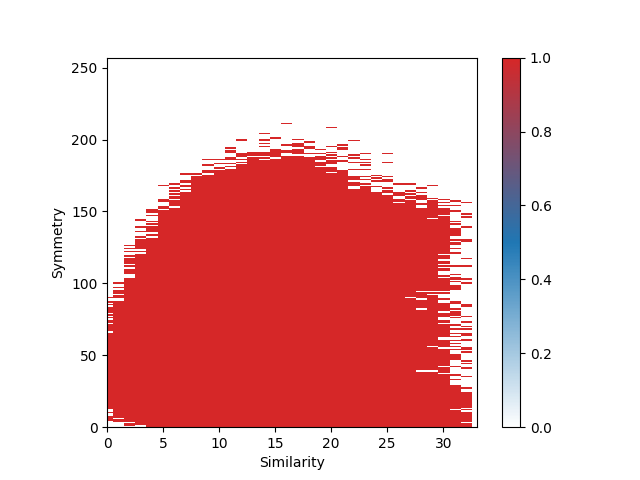}
\caption{SMB-KI-MM}
\end{subfigure}
\caption{\label{XFIGUREsymsim} Archives for the Symmetry-Similarity BC.}
\end{figure*}
}

\newcommand{\XFIGUREdenllevels}{
\begin{figure*}[t!]
\centering
\begin{subfigure}[t]{0.2\textwidth}
\centering
\raisebox{30pt}{\rotatebox{90}{\large{\textbf{SMB}}}}
\includegraphics[width=0.75\linewidth]{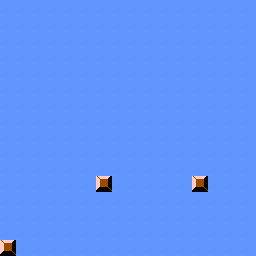}
\caption{Den=3, NL=2} 
\end{subfigure}
~
\begin{subfigure}[t]{0.2\textwidth}
\centering
\includegraphics[width=0.75\linewidth]{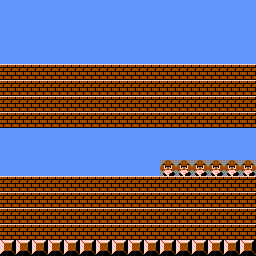}
\caption{Den=150, NL=0} 
\end{subfigure}
~
\begin{subfigure}[t]{0.2\textwidth}
\centering
\includegraphics[width=0.75\linewidth]{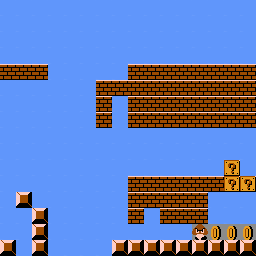}
\caption{Den=75 NL=15}
\end{subfigure}
~
\begin{subfigure}[t]{0.2\textwidth}
\centering
\includegraphics[width=0.75\linewidth]{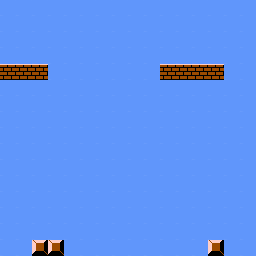}
\caption{Den=10, NL=30}
\end{subfigure}
~
\begin{subfigure}[t]{0.2\textwidth}
\centering
\includegraphics[width=0.75\linewidth]{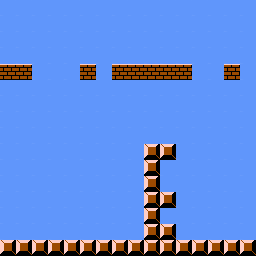}
\caption{Den=34, NL=30}
\end{subfigure}
\newline
\begin{subfigure}[t]{0.2\textwidth}
\centering
\raisebox{30pt}{\rotatebox{90}{\large{\textbf{KI}}}}
\includegraphics[width=0.75\linewidth]{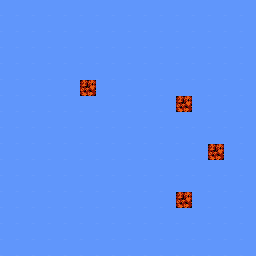}
\caption{Den=4, NL=11} 
\end{subfigure}
~
\begin{subfigure}[t]{0.2\textwidth}
\centering
\includegraphics[width=0.75\linewidth]{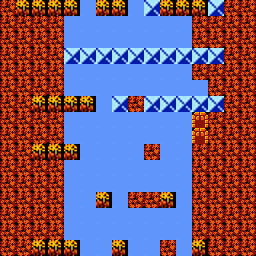}
\caption{Den=157, NL=1} 
\end{subfigure}
~
\begin{subfigure}[t]{0.2\textwidth}
\centering
\includegraphics[width=0.75\linewidth]{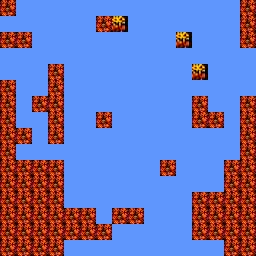}
\caption{Den=80, NL=26}
\end{subfigure}
~
\begin{subfigure}[t]{0.2\textwidth}
\centering
\includegraphics[width=0.75\linewidth]{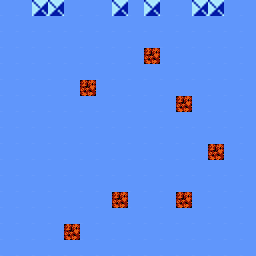}
\caption{Den=13, NL=48}
\end{subfigure}
~
\begin{subfigure}[t]{0.2\textwidth}
\centering
\includegraphics[width=0.75\linewidth]{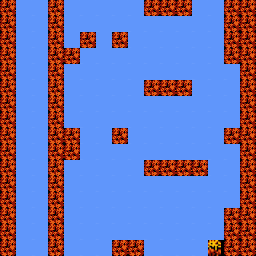}
\caption{Den=75, NL=43}
\end{subfigure}
\newline
\begin{subfigure}[t]{0.2\textwidth}
\centering

\raisebox{30pt}{\rotatebox{90}{\large{\textbf{MM}}}}
\includegraphics[width=0.75\linewidth]{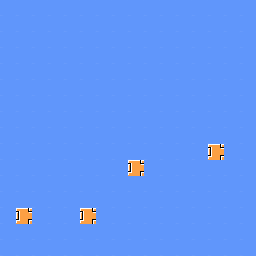}
\caption{Den=4, NL=3}
\end{subfigure}
~
\begin{subfigure}[t]{0.2\textwidth}
\centering
\includegraphics[width=0.75\linewidth]{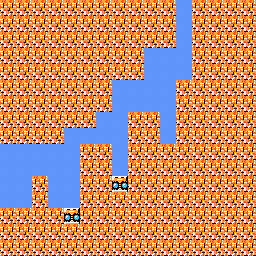}
\caption{Den=211, NL=1}
\end{subfigure}
~
\begin{subfigure}[t]{0.2\textwidth}
\centering
\includegraphics[width=0.75\linewidth]{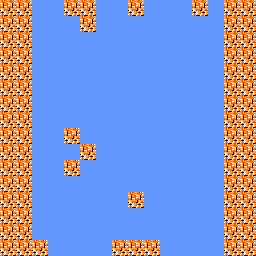}
\caption{Den=77, NL=56}
\end{subfigure}
~
\begin{subfigure}[t]{0.2\textwidth}
\centering
\includegraphics[width=0.75\linewidth]{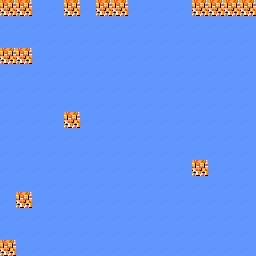}
\caption{Den=15, NL=55}
\end{subfigure}
~
\begin{subfigure}[t]{0.2\textwidth}
\centering
\includegraphics[width=0.75\linewidth]{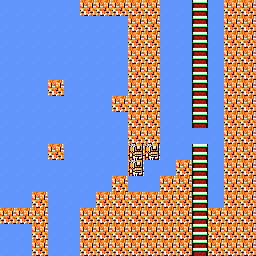}
\caption{Den=114, NL=30}
\end{subfigure}
\newline
\begin{subfigure}[t]{0.2\textwidth}
\centering
\raisebox{30pt}{\rotatebox{90}{\large{\textbf{CV}}}}
\includegraphics[width=0.75\linewidth]{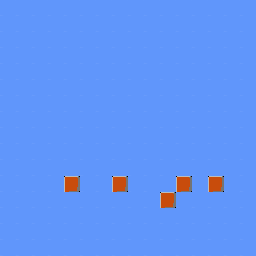}
\caption{Den=5, NL=3}
\end{subfigure}
~
\begin{subfigure}[t]{0.2\textwidth}
\centering
\includegraphics[width=0.75\linewidth]{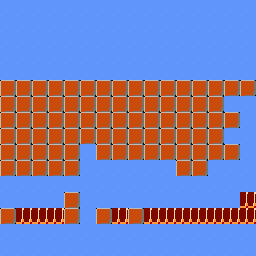}
\caption{Den=97, NL=0}
\end{subfigure}
~
\begin{subfigure}[t]{0.2\textwidth}
\centering
\includegraphics[width=0.75\linewidth]{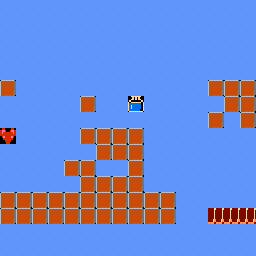}
\caption{Den=50, NL=12}
\end{subfigure}
~
\begin{subfigure}[t]{0.2\textwidth}
\centering
\includegraphics[width=0.75\linewidth]{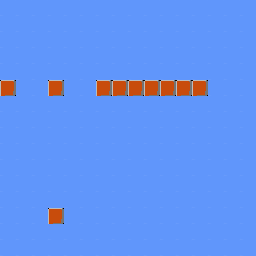}
\caption{Den=10, NL=25}
\end{subfigure}
~
\begin{subfigure}[t]{0.2\textwidth}
\centering
\includegraphics[width=0.75\linewidth]{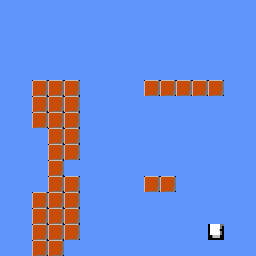}
\caption{Den=35, NL=25}
\end{subfigure}
\newline
\begin{subfigure}[t]{0.2\textwidth}
\centering
\raisebox{30pt}{\rotatebox{90}{\large{\textbf{NG}}}}
\includegraphics[width=0.75\linewidth]{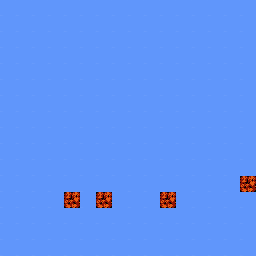}
\caption{Den=4, NL=2}
\end{subfigure}
~
\begin{subfigure}[t]{0.2\textwidth}
\centering
\includegraphics[width=0.75\linewidth]{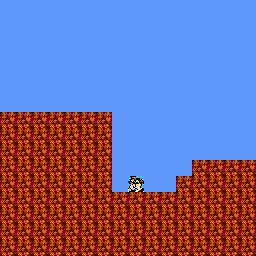}
\caption{Den=109, NL=2}
\end{subfigure}
~
\begin{subfigure}[t]{0.2\textwidth}
\centering
\includegraphics[width=0.75\linewidth]{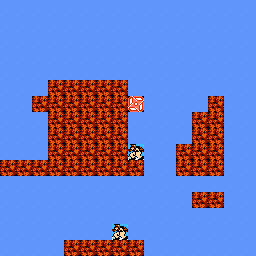}
\caption{Den=56, NL=13}
\end{subfigure}
~
\begin{subfigure}[t]{0.2\textwidth}
\centering
\includegraphics[width=0.75\linewidth]{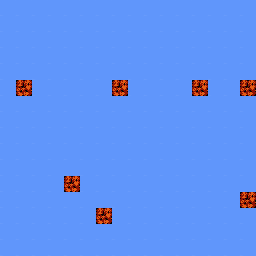}
\caption{Den=7, NL=18}
\end{subfigure}
~
\begin{subfigure}[t]{0.2\textwidth}
\centering
\includegraphics[width=0.75\linewidth]{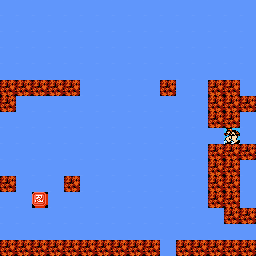}
\caption{Den=44, NL=23}
\end{subfigure}
\newline
\begin{subfigure}[t]{0.2\textwidth}
\centering
\raisebox{10pt}{\rotatebox{90}{\large{\textbf{Blend-Elites}}}}
\includegraphics[width=0.75\linewidth]{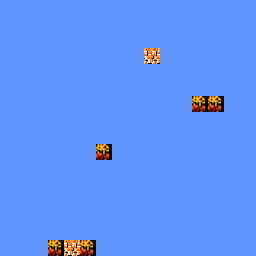}
\caption{Den=7, NL=15}
\end{subfigure}
~
\begin{subfigure}[t]{0.2\textwidth}
\centering
\includegraphics[width=0.75\linewidth]{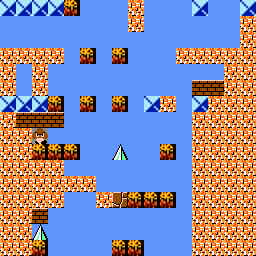}
\caption{Den=124, NL=10}
\end{subfigure}
~
\begin{subfigure}[t]{0.2\textwidth}
\centering
\includegraphics[width=0.75\linewidth]{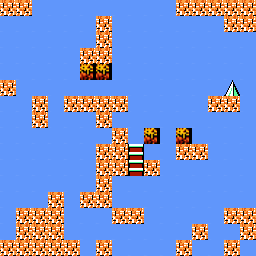}
\caption{Den=62, NL=22}
\end{subfigure}
~
\begin{subfigure}[t]{0.2\textwidth}
\centering
\includegraphics[width=0.75\linewidth]{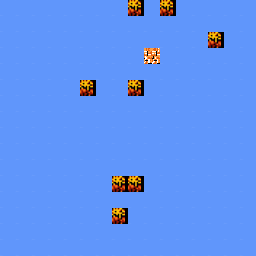}
\caption{Den=9, NL=34}
\end{subfigure}
~
\begin{subfigure}[t]{0.2\textwidth}
\centering
\includegraphics[width=0.75\linewidth]{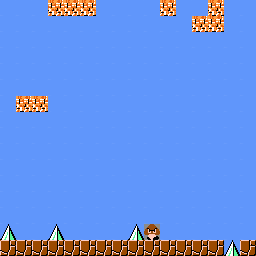}
\caption{Den=29, NL=44}
\end{subfigure}
\caption{\label{XFIGUREdenllevels} Example generated levels under the Density-Nonlinearity BC for each game. Each example is an elite from the archive.}
\end{figure*}
\vspace*{5cm}
}

\newcommand{\XFIGUREsymsimlevels}{
\begin{figure*}[t!]
\centering
\begin{subfigure}[t]{0.2\textwidth}
\centering
\raisebox{30pt}{\rotatebox{90}{\large{\textbf{SMB}}}}
\includegraphics[width=0.75\linewidth]{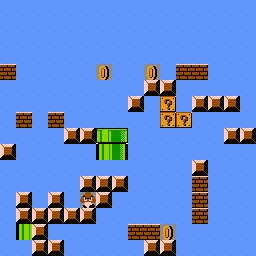}
\caption{Sym=2, Sim=2} 
\end{subfigure}
~
\begin{subfigure}[t]{0.2\textwidth}
\centering
\includegraphics[width=0.75\linewidth]{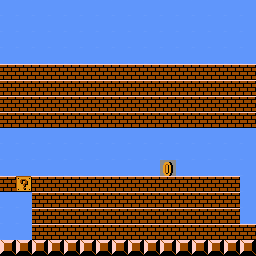}
\caption{Sym=102, Sim=15} 
\end{subfigure}
~
\begin{subfigure}[t]{0.2\textwidth}
\centering
\includegraphics[width=0.75\linewidth]{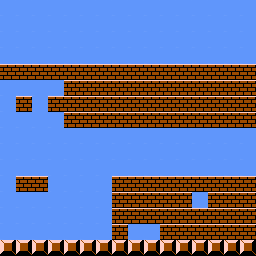}
\caption{Sym=60 Sim=17}
\end{subfigure}
~
\begin{subfigure}[t]{0.2\textwidth}
\centering
\includegraphics[width=0.75\linewidth]{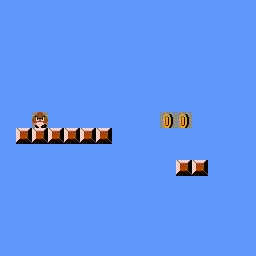}
\caption{Sym=0, Sim=32}
\end{subfigure}
~
\begin{subfigure}[t]{0.2\textwidth}
\centering
\includegraphics[width=0.75\linewidth]{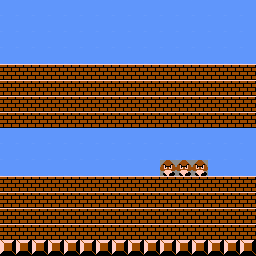}
\caption{Sym=120, Sim=32}
\end{subfigure}
\newline
\begin{subfigure}[t]{0.2\textwidth}
\centering
\raisebox{30pt}{\rotatebox{90}{\large{\textbf{KI}}}}
\includegraphics[width=0.75\linewidth]{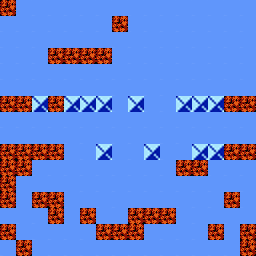}
\caption{Sym=6, Sim=0} 
\end{subfigure}
~
\begin{subfigure}[t]{0.2\textwidth}
\centering
\includegraphics[width=0.75\linewidth]{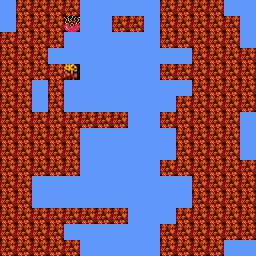}
\caption{Sym=91, Sim=5} 
\end{subfigure}
~
\begin{subfigure}[t]{0.2\textwidth}
\centering
\includegraphics[width=0.75\linewidth]{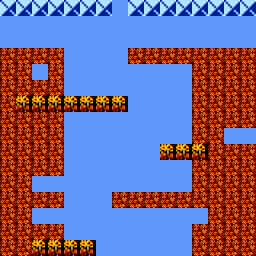}
\caption{Sym=45, Sim=16}
\end{subfigure}
~
\begin{subfigure}[t]{0.2\textwidth}
\centering
\includegraphics[width=0.75\linewidth]{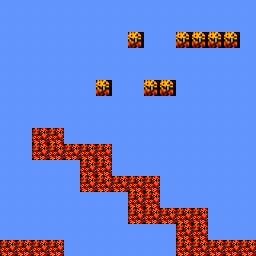}
\caption{Sym=1, Sim=32}
\end{subfigure}
~
\begin{subfigure}[t]{0.2\textwidth}
\centering
\includegraphics[width=0.75\linewidth]{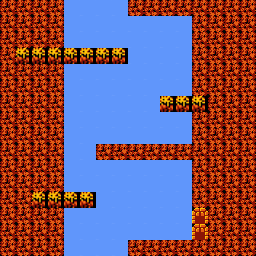}
\caption{Sym=71, sim=32}
\end{subfigure}
\newline
\begin{subfigure}[t]{0.2\textwidth}
\centering
\raisebox{30pt}{\rotatebox{90}{\large{\textbf{MM}}}}
\includegraphics[width=0.75\linewidth]{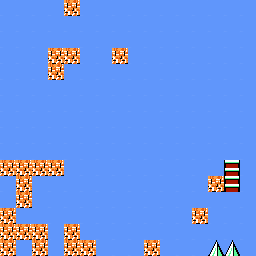}
\caption{Sym=0, Sim=10}
\end{subfigure}
~
\begin{subfigure}[t]{0.2\textwidth}
\centering
\includegraphics[width=0.75\linewidth]{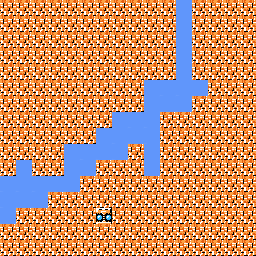}
\caption{Sym=181, Sim=20}
\end{subfigure}
~
\begin{subfigure}[t]{0.2\textwidth}
\centering
\includegraphics[width=0.75\linewidth]{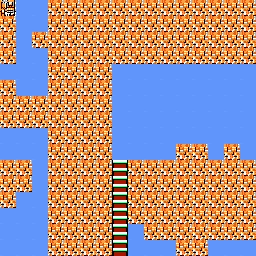}
\caption{Sym=107, Sim=19}
\end{subfigure}
~
\begin{subfigure}[t]{0.2\textwidth}
\centering
\includegraphics[width=0.75\linewidth]{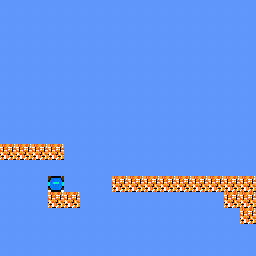}
\caption{Sym=1, Sim=32}
\end{subfigure}
~
\begin{subfigure}[t]{0.2\textwidth}
\centering
\includegraphics[width=0.75\linewidth]{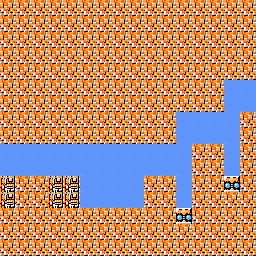}
\caption{Sym=159, Sim=32}
\end{subfigure}
\newline
\begin{subfigure}[t]{0.2\textwidth}
\centering
\raisebox{30pt}{\rotatebox{90}{\large{\textbf{CV}}}}
\includegraphics[width=0.75\linewidth]{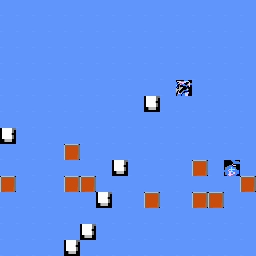}
\caption{Sym=0, Sim=8}
\end{subfigure}
~
\begin{subfigure}[t]{0.2\textwidth}
\centering
\includegraphics[width=0.75\linewidth]{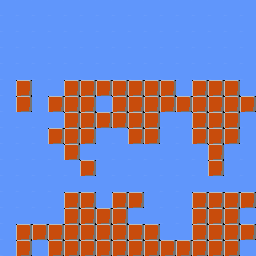}
\caption{Sym=58, Sim=12}
\end{subfigure}
~
\begin{subfigure}[t]{0.2\textwidth}
\centering
\includegraphics[width=0.75\linewidth]{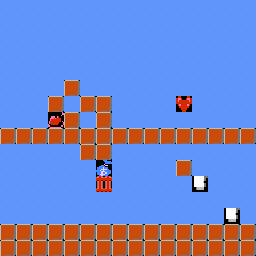}
\caption{Sym=29, Sim=18}
\end{subfigure}
~
\begin{subfigure}[t]{0.2\textwidth}
\centering
\includegraphics[width=0.75\linewidth]{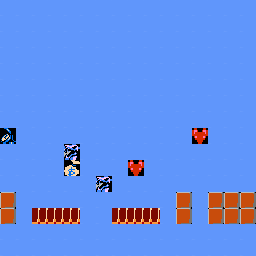}
\caption{Sym=0, Sim=32}
\end{subfigure}
~
\begin{subfigure}[t]{0.2\textwidth}
\centering
\includegraphics[width=0.75\linewidth]{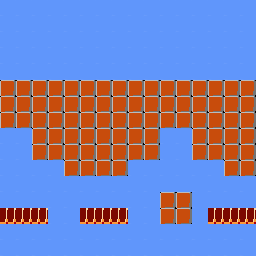}
\caption{Sym=39, Sim=32}
\end{subfigure}
\newline
\begin{subfigure}[t]{0.2\textwidth}
\centering
\raisebox{30pt}{\rotatebox{90}{\large{\textbf{NG}}}}
\includegraphics[width=0.75\linewidth]{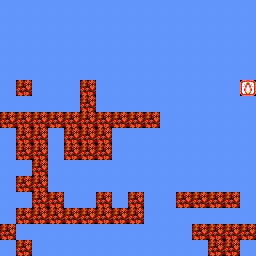}
\caption{Sym=7, Sim=5}
\end{subfigure}
~
\begin{subfigure}[t]{0.2\textwidth}
\centering
\includegraphics[width=0.75\linewidth]{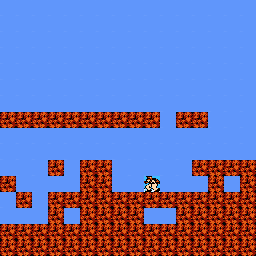}
\caption{Sym=45, Sim=11}
\end{subfigure}
~
\begin{subfigure}[t]{0.2\textwidth}
\centering
\includegraphics[width=0.75\linewidth]{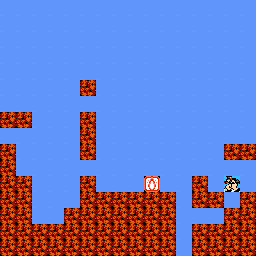}
\caption{Sym=26, Sim=17}
\end{subfigure}
~
\begin{subfigure}[t]{0.2\textwidth}
\centering
\includegraphics[width=0.75\linewidth]{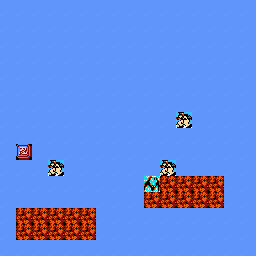}
\caption{Sym=0, Sim=32}
\end{subfigure}
~
\begin{subfigure}[t]{0.2\textwidth}
\centering
\includegraphics[width=0.75\linewidth]{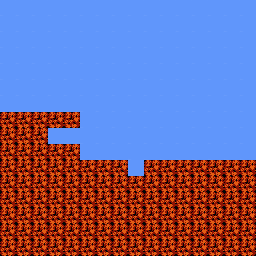}
\caption{Sym=52, Sim=27}
\end{subfigure}
\newline
\begin{subfigure}[t]{0.2\textwidth}
\centering
\raisebox{10pt}{\rotatebox{90}{\large{\textbf{Blend-Elites}}}}
\includegraphics[width=0.75\linewidth]{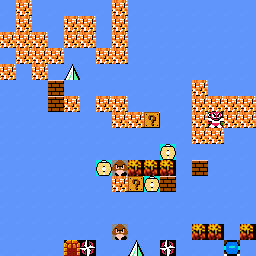}
\caption{Sym=5, Sim=0}
\end{subfigure}
~
\begin{subfigure}[t]{0.2\textwidth}
\centering
\includegraphics[width=0.75\linewidth]{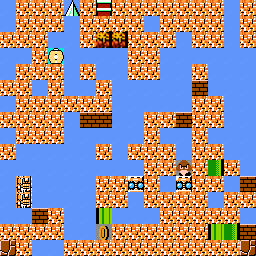}
\caption{Sym=57, Sim=0}
\end{subfigure}
~
\begin{subfigure}[t]{0.2\textwidth}
\centering
\includegraphics[width=0.75\linewidth]{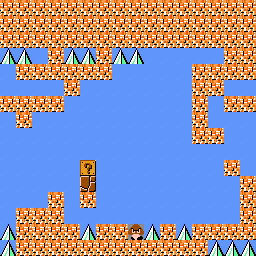}
\caption{Sym=53, Sim=5}
\end{subfigure}
~
\begin{subfigure}[t]{0.2\textwidth}
\centering
\includegraphics[width=0.75\linewidth]{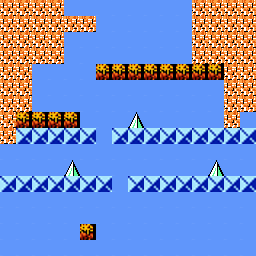}
\caption{Sym=16, Sim=28}
\end{subfigure}
~
\begin{subfigure}[t]{0.2\textwidth}
\centering
\includegraphics[width=0.75\linewidth]{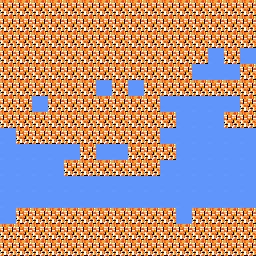}
\caption{Sym=126, Sim=11}
\end{subfigure}
\caption{\label{XFIGUREsymsimlevels} Example generated levels under the Symmetry-Similarity BC for each game. Each example is an elite from the archive.}
\end{figure*}
\vspace*{10cm}
}

\newcommand{\XFIGUREelemlevels}{
\begin{figure*}[t!]
\centering
\begin{subfigure}[t]{0.2\textwidth}
\centering
\raisebox{30pt}{\rotatebox{90}{\large{\textbf{SMB}}}}
\includegraphics[width=0.75\linewidth]{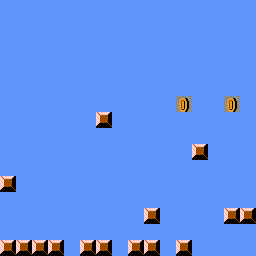}
\caption{00010} 
\end{subfigure}
~
\begin{subfigure}[t]{0.2\textwidth}
\centering
\includegraphics[width=0.75\linewidth]{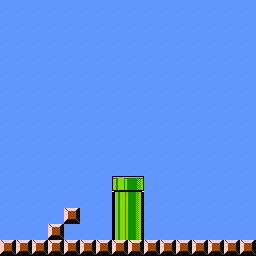}
\caption{01000}
\end{subfigure}
~
\begin{subfigure}[t]{0.2\textwidth}
\centering
\includegraphics[width=0.75\linewidth]{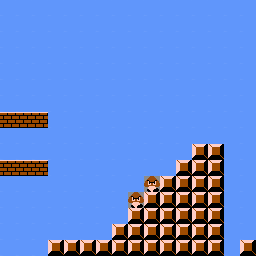}
\caption{10001}
\end{subfigure}
~
\begin{subfigure}[t]{0.2\textwidth}
\centering
\includegraphics[width=0.75\linewidth]{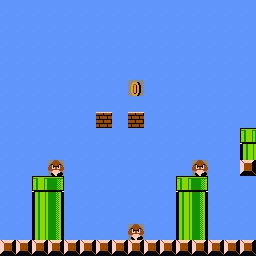}
\caption{11011}
\end{subfigure}
~
\begin{subfigure}[t]{0.2\textwidth}
\centering
\includegraphics[width=0.75\linewidth]{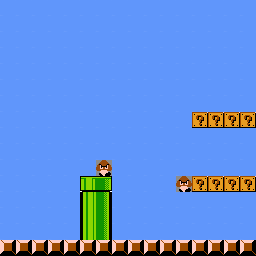}
\caption{11100} 
\end{subfigure}
\newline
\begin{subfigure}[t]{0.2\textwidth}
\centering
\raisebox{30pt}{\rotatebox{90}{\large{\textbf{KI}}}}
\includegraphics[width=0.75\linewidth]{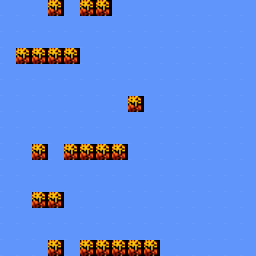}
\caption{0001} 
\end{subfigure}
~
\begin{subfigure}[t]{0.2\textwidth}
\centering
\includegraphics[width=0.75\linewidth]{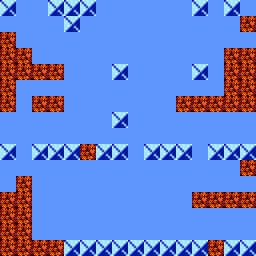}
\caption{0010} 
\end{subfigure}
~
\begin{subfigure}[t]{0.2\textwidth}
\centering
\includegraphics[width=0.75\linewidth]{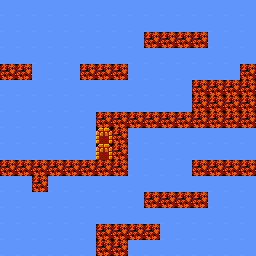}
\caption{0100}
\end{subfigure}
~
\begin{subfigure}[t]{0.2\textwidth}
\centering
\includegraphics[width=0.75\linewidth]{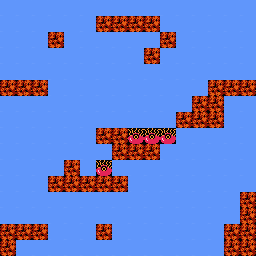}
\caption{1000}
\end{subfigure}
~
\begin{subfigure}[t]{0.2\textwidth}
\centering
\includegraphics[width=0.75\linewidth]{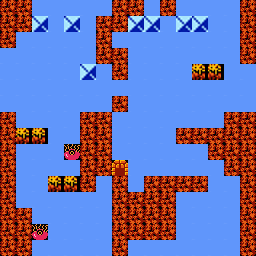}
\caption{1111}
\end{subfigure}
\newline
\begin{subfigure}[t]{0.2\textwidth}
\centering
\raisebox{30pt}{\rotatebox{90}{\large{\textbf{MM}}}}
\includegraphics[width=0.75\linewidth]{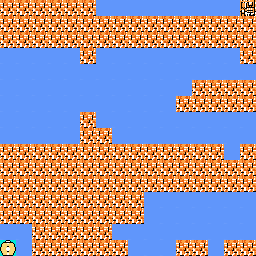}
\caption{00001}
\end{subfigure}
~
\begin{subfigure}[t]{0.2\textwidth}
\centering
\includegraphics[width=0.75\linewidth]{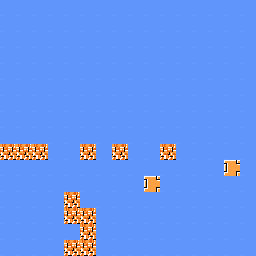}
\caption{00010}
\end{subfigure}
~
\begin{subfigure}[t]{0.2\textwidth}
\centering
\includegraphics[width=0.75\linewidth]{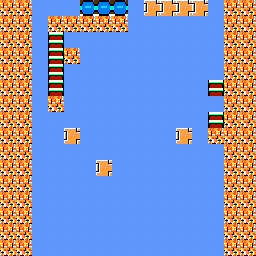}
\caption{00111}
\end{subfigure}
~
\begin{subfigure}[t]{0.2\textwidth}
\centering
\includegraphics[width=0.75\linewidth]{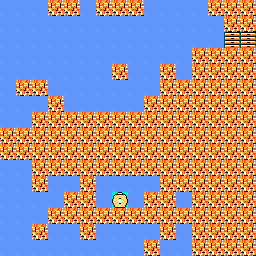}
\caption{01001}
\end{subfigure}
~
\begin{subfigure}[t]{0.2\textwidth}
\centering
\includegraphics[width=0.75\linewidth]{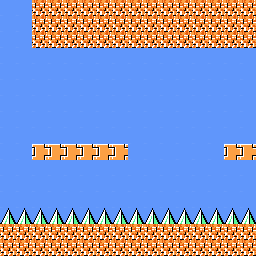}
\caption{10010}
\end{subfigure}
\newline
\begin{subfigure}[t]{0.2\textwidth}
\centering
\raisebox{30pt}{\rotatebox{90}{\large{\textbf{CV}}}}
\includegraphics[width=0.75\linewidth]{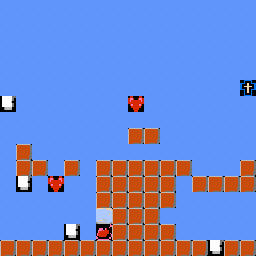}
\caption{0011101}
\end{subfigure}
~
\begin{subfigure}[t]{0.2\textwidth}
\centering
\includegraphics[width=0.75\linewidth]{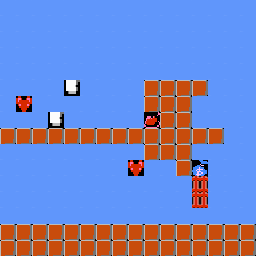}
\caption{0110100}
\end{subfigure}
~
\begin{subfigure}[t]{0.2\textwidth}
\centering
\includegraphics[width=0.75\linewidth]{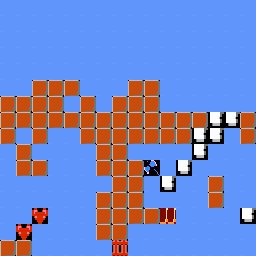}
\caption{0111110}
\end{subfigure}
~
\begin{subfigure}[t]{0.2\textwidth}
\centering
\includegraphics[width=0.75\linewidth]{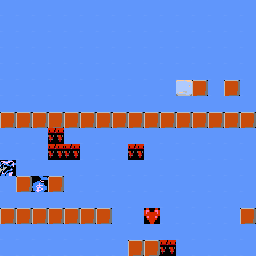}
\caption{1000101}
\end{subfigure}
~
\begin{subfigure}[t]{0.2\textwidth}
\centering
\includegraphics[width=0.75\linewidth]{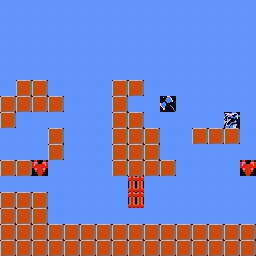}
\caption{1101100}
\end{subfigure}
\newline
\begin{subfigure}[t]{0.2\textwidth}
\centering
\raisebox{30pt}{\rotatebox{90}{\large{\textbf{NG}}}}
\includegraphics[width=0.75\linewidth]{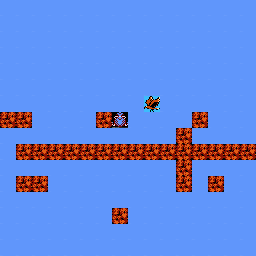}
\caption{01001}
\end{subfigure}
~
\begin{subfigure}[t]{0.2\textwidth}
\centering
\includegraphics[width=0.75\linewidth]{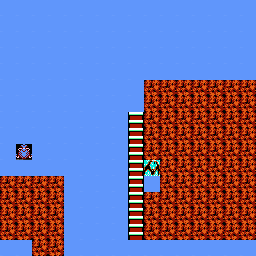}
\caption{01101}
\end{subfigure}
~
\begin{subfigure}[t]{0.2\textwidth}
\centering
\includegraphics[width=0.75\linewidth]{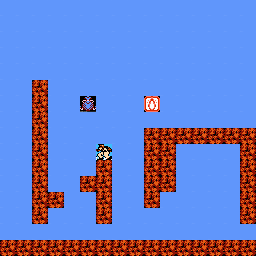}
\caption{10011}
\end{subfigure}
~
\begin{subfigure}[t]{0.2\textwidth}
\centering
\includegraphics[width=0.75\linewidth]{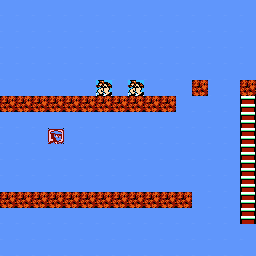}
\caption{10101}
\end{subfigure}
~
\begin{subfigure}[t]{0.2\textwidth}
\centering
\includegraphics[width=0.75\linewidth]{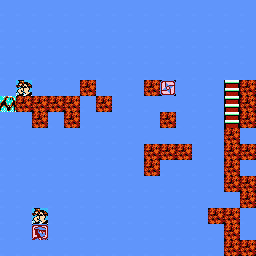}
\caption{11111}
\end{subfigure}
\newline
\begin{subfigure}[t]{0.2\textwidth}
\centering
\raisebox{10pt}{\rotatebox{90}{\large{\textbf{Blend-Elites}}}}
\includegraphics[width=0.75\linewidth]{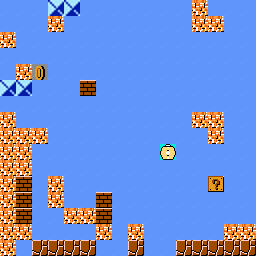}
\caption{000011101}
\end{subfigure}
~
\begin{subfigure}[t]{0.2\textwidth}
\centering
\includegraphics[width=0.75\linewidth]{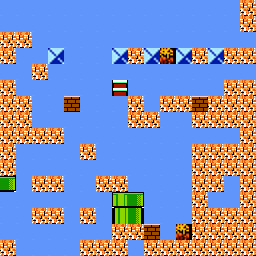}
\caption{001100111}
\end{subfigure}
~
\begin{subfigure}[t]{0.2\textwidth}
\centering
\includegraphics[width=0.75\linewidth]{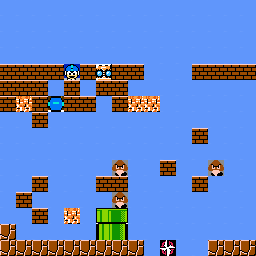}
\caption{100101011}
\end{subfigure}
~
\begin{subfigure}[t]{0.2\textwidth}
\centering
\includegraphics[width=0.75\linewidth]{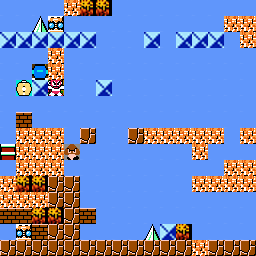}
\caption{101001111}
\end{subfigure}
~
\begin{subfigure}[t]{0.2\textwidth}
\centering
\includegraphics[width=0.75\linewidth]{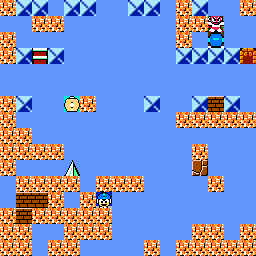}
\caption{111001101}
\end{subfigure}
\caption{\label{XFIGUREelemlevels} Example generated levels under the Game Element BC for each game. Each example is an elite from the archive. Labels indicate: \textit{\textbf{SMB:} Enemy, Pipe, ?-Mark, Collectable, Breakable}, \textit{\textbf{KI:} Hazard, Door, Moving Platform, Fixed Platform}, \textit{\textbf{MM:} Enemy/Hazard, Ladder, Platform, Collectable}, \textit{\textbf{CV:} Enemy, Door, Ladder, Weapon, Collectable, Moving, Breakable}, \textit{\textbf{NG:} Enemy, Animal, Ladder, Weapon, Collectable}, \textit{\textbf{Blend-Elites:} Enemy, Door, Ladder, Pipe, ?-Mark, Collectable, Moving Platform, Fixed Platform, Breakable}}
\end{figure*}
}

\newcommand{\XFIGUREblendplots}{
\begin{figure*}[t!]
\centering
\begin{subfigure}[t]{0.45\textwidth}
\centering
\includegraphics[height=2.3in]{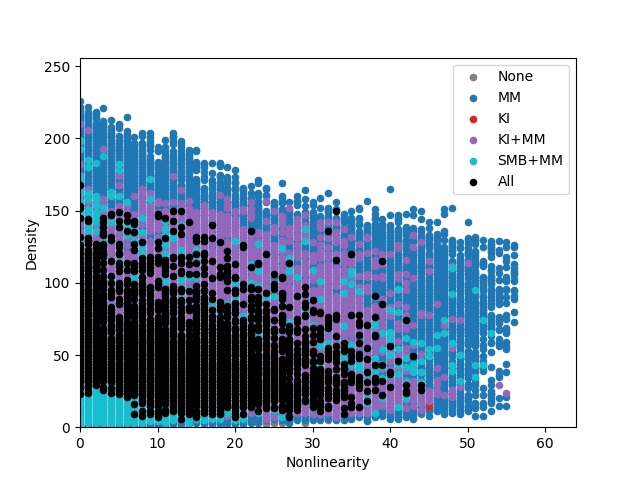}
\caption{Density-Nonlinearity}
\end{subfigure}
\begin{subfigure}[t]{0.45\textwidth}
\centering
\includegraphics[height=2.3in]{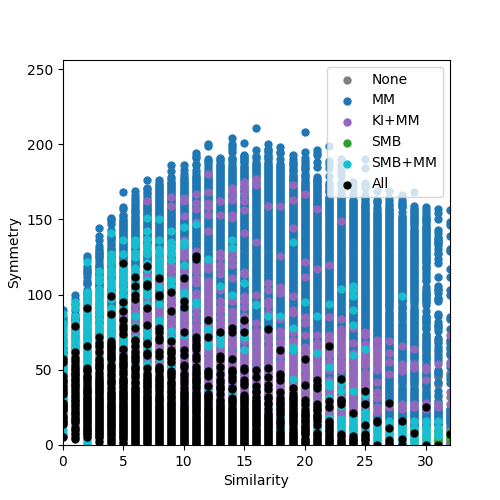}
\caption{Symmetry-Similarity}
\end{subfigure}
\caption{\label{XFIGUREblendplots} Archive of tile-based BCs for \textit{Blend-Elites} with each cell colored based on the set of agents that was able to complete a segment assigned to that cell.}
\end{figure*}
}

\newcommand{\XFIGUREdenlplots}{
\begin{figure*}[t!]
\centering
\begin{subfigure}[t]{0.5\textwidth}
\centering
\includegraphics[width=0.6\linewidth]{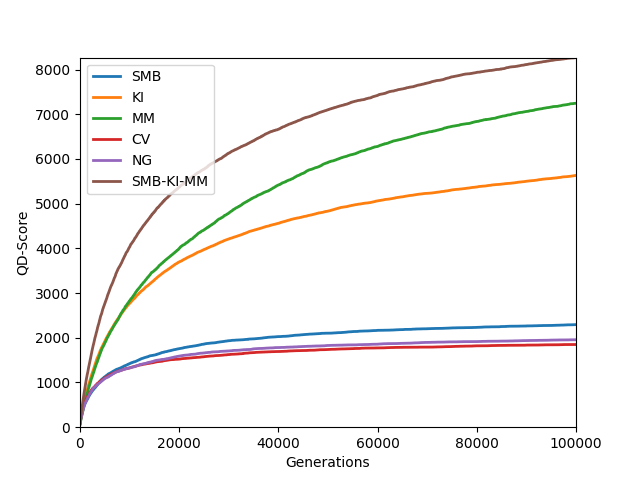}
\caption{QD-Score}
\end{subfigure}
~
\begin{subfigure}[t]{0.5\textwidth}
\centering
\includegraphics[width=0.6\linewidth]{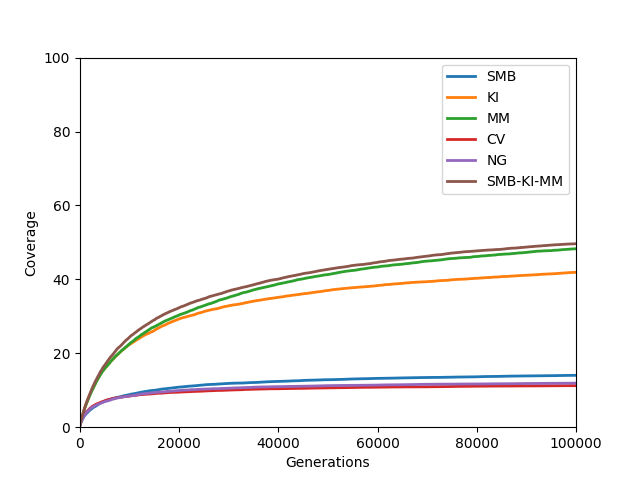}
\caption{Coverage}
\end{subfigure}
\caption{\label{XFIGUREdenlplots} QD-Scores and Coverage percentages over time for each game for the Density-Nonlinearity BC.}
\end{figure*}
}

\newcommand{\XFIGUREsymsimplots}{
\begin{figure*}[t!]
\centering
\begin{subfigure}[t]{0.5\textwidth}
\centering
\includegraphics[width=0.6\linewidth]{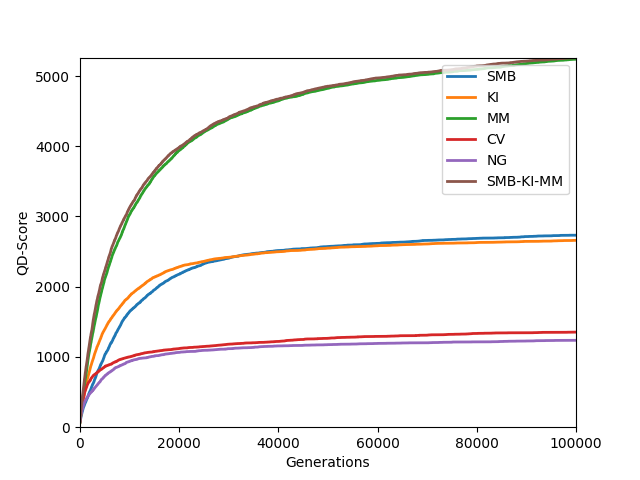}
\caption{QD-Score}
\end{subfigure}
~
\begin{subfigure}[t]{0.5\textwidth}
\centering
\includegraphics[width=0.6\linewidth]{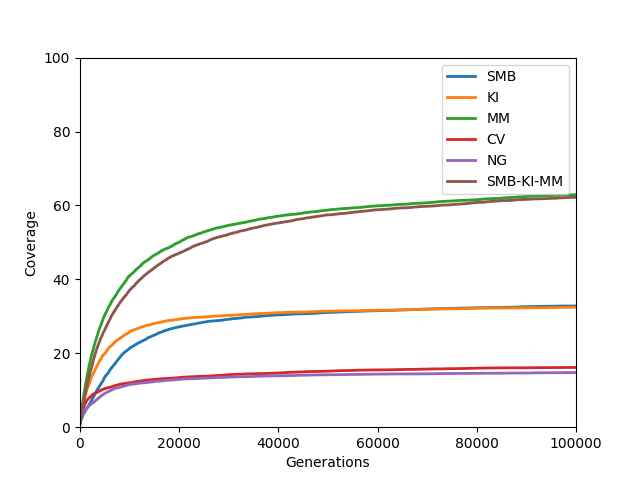}
\caption{Coverage}
\end{subfigure}
\caption{\label{XFIGUREsymsimplots} QD-Scores and Coverage percentages over time for each game for the Symmetry-Similarity BC.}
\end{figure*}
}

\newcommand{\XFIGUREelemplots}{
\begin{figure*}[t!]
\centering
\begin{subfigure}[t]{0.5\textwidth}
\centering
\includegraphics[width=0.6\linewidth]{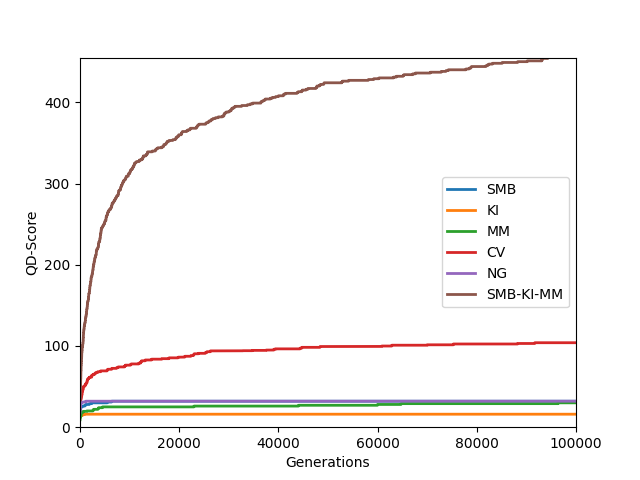}
\caption{QD-Score}
\end{subfigure}
~
\begin{subfigure}[t]{0.5\textwidth}
\centering
\includegraphics[width=0.6\linewidth]{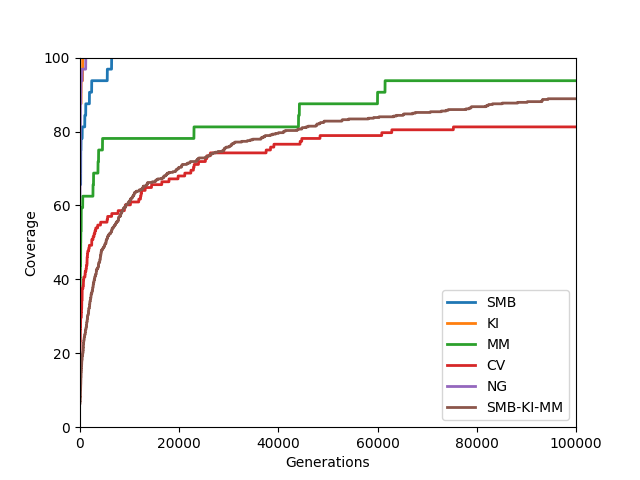}
\caption{Coverage}
\end{subfigure}
\caption{\label{XFIGUREelemplots} QD-Scores and Coverage percentages over time for each game for the Game Elements BC. Due to varying archive sizes across games, some games converge a lot faster than others.}
\end{figure*}
}


\newcommand{\XTABLEqdcov}{\begin{table*}[t!]
\centering
\small
\begin{tabular}{|c|c|c|c||c|c|c||c|c|c|}
\hline
\multirow{2}{*}{} & \multicolumn{3}{c||}{Density-Nonlinearity} & \multicolumn{3}{c||}{Symmetry-Similarity} & \multicolumn{3}{c|}{Game-Elements} \\ \cline{2-10} 
 & QD-Score & Coverage & \% Optimal & QD-Score & Coverage & \% Optimal & QD-Score & Coverage & \% Optimal\\ \hline
SMB &  2341.81 & 14.03  & 97.48 & 2726.00 & 32.14 & 98.97 & \hphantom{0}32.00 & 100.00 & 100  \\ \hline
KI & 5631.81 & 41.91 & 67.4 & 2660.31 & 32.48 & 94.22 &  \hphantom{0}16.00  & 100.00 & 100 \\ \hline
MM & 7245.94 & 48.29 & 89.26 & 5239.06 & 62.9 & 97.75 & \hphantom{0}30  & \hphantom{0}93.75 & 100 \\ \hline
CV & 1849.38 & 11.24  & 97.97 & 1353.63 & 16.17 & 97.81 & 104.00 & \hphantom{0}81.25 & 100\\ \hline
NG & 1955.94 & 11.92 & 97.69 & 1237.13  & 14.8 & 98.41 & \hphantom{0}32.00 & 100.00 & 100\\ \hline
Blend-Elites &  8267.31 & 49.64  & 99.66 & 5262 & 62.22 & 99.72 & 455.00 & \hphantom{0}88.87 & 100\\ \hline
\end{tabular}
\caption{\label{XTABLEqdcov} For all experiments, QD-Score and Coverage values along with percentage of covered cells that had optimal playability.}
\end{table*}
}

\newcommand{\XTABLEqdcovless}{\begin{table*}[t!]
\centering
\begin{tabular}{|c|c|c||c|c||c|c|}
\hline
\multirow{2}{*}{} & \multicolumn{2}{c||}{Density-Nonlinearity (s=16705)} & \multicolumn{2}{c||}{Symmetry-Similarity (s=8481)} & \multicolumn{2}{c|}{Game-Elements} \\ \cline{2-7} 
 & QD-Score & Coverage & QD-Score & Coverage & QD-Score & Coverage\\ \hline
SMB &  2341.81 & 14.03  & 2726.00 & 32.14 &  \hphantom{0}32.00 & 100.00\\ \hline
KI & 5631.81 & 41.91 & 2660.31 & 32.48 &  \hphantom{0}16.00  & 100.00 \\ \hline
MM & 7245.94 & 48.29 & 5239.06 & 62.9 & \hphantom{0}30  & \hphantom{0}93.75\\ \hline
CV & 1849.38 & 11.24  & 1353.63 & 16.17 & 104.00 & \hphantom{0}81.25\\ \hline
NG & 1955.94 & 11.92 & 1237.13  & 14.8  & \hphantom{0}32.00 & 100.00\\ \hline
Blend &  8267.31 & 49.64  & 5262 & 62.22 & 455.00 & \hphantom{0}88.87\\ \hline
\end{tabular}
\caption{\label{XTABLEqdcov} For all experiments, QD-Score and Coverage values along with size of the respective archives.}
\end{table*}
}
\section{Introduction}
In recent years, variational autoencoders (VAEs) \cite{kingma2013autoencoding} have been increasingly used as a means of generating game levels as well as blending levels across games. VAEs consist of encoder-decoder neural networks and learn a continuous, lower-dimensional, latent representation of the levels used for training. This learned latent space is then sampled to generate new levels in the style of the levels in the training data. However, while shown to be capable of generating levels in the style of specific games as well as in styles that blend multiple games together, standard VAEs are not amenable to producing a diverse range of desired content in a controllable manner, since levels are generated via random sampling. Methods such as conditional VAEs \cite{sarkar2020conditional} and latent variable evolution \cite{bontrager2018deepmasterprints} have been shown to add controllability to these models but neither is well suited to produce a wide variety of content. Conditional VAEs enable the use of labels to specify desired properties of generated levels and afford controllability as a byproduct of the training process, thus avoiding having to run evolutionary search post-training. However, this comes at the cost of requiring the set of possible labels to be defined at training time. Additionally, levels are still sampled at random and controllability is achieved by modifying random vectors via labels rather than by exploring the search space. This latter approach is employed by latent variable evolution (LVE) which refers to the process of capturing a desired property using an objective function and then searching the latent space of the model using an evolutionary search algorithm such as CMA-ES \cite{hansen2003reducing} to find latent vectors that optimize this objective. LVE has shown to be effective both with VAEs \cite{sarkar2019blending} and with Generative Adversarial Networks (GANs) \cite{volz2018evolving} but is by design meant to focus on finding a single optimal level that satisfies the objective rather than finding a diverse array of levels. Additionally, this requires that the desirable properties of a level can be adequately framed as a single objective function to be optimized, which may not always be the case. Further, since this approach works with only one objective at a time, searching for different types of content necessitates multiple separate runs and/or constructing complex objective functions. Moreover, it is quite possible that the algorithm always converges to the same or similar optimal solution in each run. On the other hand, quality-diversity (QD) \cite{pugh2016quality} methods are explicitly designed to produce a range of diverse content in one evolutionary run. MAP-Elites \cite{mouret2015illuminating} and its variants have been used for generating a diverse range of game content \cite{fontaine2019mapping,khalifa2018talakat}. Hence, applying MAP-Elites could improve existing applications of VAEs by allowing them to generate and blend more diverse content. 

In this work, we present a hybrid Procedural Content Generation via Machine Learning (PCGML) approach that combines the use of VAEs with the MAP-Elites algorithm. We train a VAE each on levels from the platformers \textit{Super Mario Bros.}, \textit{Kid Icarus}, \textit{Mega Man}, \textit{Castlevania} and \textit{Ninja Gaiden} as well as one trained on the blended Mario-Icarus-Mega Man domain, \review{an approach we term \textit{Blend-Elites}}. For each model, we run MAP-Elites using the VAE latent space as the search space and different sets of behavior characteristics corresponding to various tile-based metrics as well as the presence of different game elements, optimizing for playability as the objective. Our experiments enable us to compare how MAP-Elites illuminates the latent spaces of different games as well as illuminate which regions of the blended latent space correspond to which games, thus demonstrating the potential of this approach for controllably generating playable levels that blend desired combinations of games, in the future. Our work thus contributes: 
\begin{enumerate}
\item a hybrid PCGML approach that combines the use of VAEs and MAP-Elites for level generation and blending, 
\item to our knowledge, the first use of MAP-Elites to generate levels of \textit{Kid Icarus}, \textit{Mega Man}, \textit{Castlevania} and \textit{Ninja Gaiden} and compare illuminated latent spaces of multiple games, 
\item \review{\textit{Blend-Elites i.e.}} the first application of MAP-Elites for blending levels across games.
\end{enumerate}

\section{Related Work}
Our work intersects two categories of approaches for procedural content generation (PCG) that have seen a significant body of work in recent years---1) PCG via Machine Learning (PCGML) \cite{summerville2017procedural} which refers to using ML techniques to build generative models by training on data from existing games and 2) PCG via Quality-Diversity (PCG-QD) \cite{gravina2019procedural} which refers to methods that build on search-based PCG \cite{togelius2011search} by using quality-diversity evolutionary algorithms \cite{pugh2016quality} for producing content. While several ML techniques such as LSTMs \cite{summerville2016mariostring}, Markov models \cite{snodgrass2017learning}, graphical models \cite{guzdial2016game} and GANs \cite{volz2018evolving} have been utilized for generating game content, most relevant to our work are variational autoencoders (VAEs). VAEs have been used to generate levels for a number of different games \cite{sarkar2019blending,thakkar2019autoencoder} as well as to produce blended levels \cite{sarkar2020exploring} and physics \cite{summerville2020extracting} that combine the properties of multiple games taken together. More advanced VAE models such as conditional VAEs and Gaussian Mixture VAEs have been used to respectively control level generation using labels \cite{sarkar2020conditional} and learn unsupervised clusters of different level types which could then be used to generate levels of that type \cite{yang2020game}. VAEs have also been used to classify NPC behaviors \cite{soares2019deep} and learn game entity embeddings \cite{khameneh2020entity}. Relatedly, several works have also focused on combining latent variable models like GANs and VAEs with evolutionary algorithms, motivated by wanting to incorporate more controllability into generation by searching the learned latent space of these models for desired content. Volz et al. \cite{volz2018evolving} demonstrated the use of running Covariance Matrix Adaptation-Evolution Strategy (CMA-ES) \cite{hansen2003reducing} for evolving desired levels using the latent space of a GAN trained on Mario levels. Schrum et al. \cite{schrum2020interactive} used an interactive evolutionary approach to allow users to evolve levels using a tool that used GAN models for level generation. Such techniques for latent variable evolution \cite{bontrager2018deepmasterprints} have also been used with VAEs for level generation and blending \cite{sarkar2019blending}. Our work builds on the latter in that we run evolutionary search within the latent space of a VAE, but use the quality-diversity algorithm MAP-Elites.

MAP-Elites \cite{mouret2015illuminating} is an evolutionary algorithm under the Quality-Diversity (QD) paradigm \cite{pugh2016quality}. QD algorithms search for a set of diverse and high-quality solutions in a single run by defining subspaces within the evolutionary search space that correspond to different behaviors and return the solution that optimizes a fitness function in each subspace. This is done by maintaining a map of cells (also referred to as archive) where each cell defines a partition or niche of the search space and stores the optimal solution for that niche. Several recent works have used variants of MAP-Elites for generating content. A constraint-based version of MAP-Elites was developed by Khalifa et al. \cite{khalifa2018talakat} for generating levels for a bullet hell game and then used by Green et al. \cite{green2018generating} for generating sections of Mario levels, by Alvarez et al. \cite{alvarez2019empowering} for evolving dungeon levels and by Charity et al. \cite{charity2020mech} to illuminate the game mechanic space of the \mbox{GVG-AI} framework and generate levels requiring specific mechanics. A variant of MAP-Elites using sliding boundaries that redefine the dimensions of the map over the course of evolution was used by Fontaine et al. \cite{fontaine2019mapping} to evolve decks for \textit{Hearthstone}. Charity et al. \cite{charity2020baba} used MAP-Elites as part of a mixed-initiative tool for designing and generating levels for the puzzle game \textit{Baba Is You}. Further, Withington \cite{withington2020illuminating} compared MAP-Elites with another QD algorithm SHINE for Mario level generation while Canaan et al. \cite{canaan2020generating} used MAP-Elites to generate diverse agents for \textit{Hanabi}. A comprehensive survey of using MAP-Elites and other QD algorithms for PCG is provided in \cite{gravina2019procedural}. Similar to these approaches, we use MAP-Elites to generate game levels but use the learned latent representation of the VAE as the evolutionary search space.

In running MAP-Elites within the latent space of an ML model, the prior work most similar to ours is that of Fontaine et al. \cite{fontaine2020illuminating} who run MAP-Elites and its variants, focusing on CMA-ME \cite{fontaine2020covariance}, a hybrid algorithm combining CMA-ES with MAP-Elites, in the latent space of a GAN trained on Mario levels, referring to the method of using QD algorithms to explore the learned latent space as \textit{latent space illumination}. Our work differs in leveraging VAEs rather than GANs, in using 4 additional games on which QD has not previously been applied and in demonstrating its use for game blending.

Other prior works have also combined ML and QD such as Innovation Engines \cite{nguyen2015innovation} and Go-Explore \cite{ecoffet2019go}. Similar to our approach, Gaier et al. \cite{gaier2020discovering} also combine the use of VAEs with MAP-Elites. However, their approach involves an iterative cycle of training a VAE on the MAP-Elites archive for representation learning and then running MAP-Elites on the learned representation in the next cycle. We instead run MAP-Elites on the latent space of the trained VAE rather than on its outputs. In the context of games, the \mbox{DeLeNox} system \cite{liapis2013transforming} used novelty search with an autoencoder to generate 2D arcade-style spaceships while Schrum et al. \cite{schrum2020cppn2gan} used MAP-Elites for latent space illumination in a hybrid method for Mario level generation that combined GANs with compositional pattern producing networks (CPPNs). In another hybrid approach, Gonzalez et al. \cite{gonzalez2020finding} used MAP-Elites in conjunction with gameplaying agents and Bayesian optimization in a process called \textit{Intelligent Trial \& Error} to generate GVG-AI levels of appropriate difficulty.

Finally, in blending levels across different games, our work follows a recent line of PCGML research focusing on more creative applications of ML for game design \cite{guzdial2018combinatorial,sarkar2020towards}. Such techniques touch upon combinational creativity \cite{boden2004creative} and have included domain transfer \cite{snodgrass2016approach,snodgrass2020multi}, automated game generation \cite{guzdial2018automated} and game blending \cite{gow2015towards} which refers to generating new games by combining the levels and/or mechanics of existing games. While many recent works have blended games via the VAE latent space \cite{sarkar2019blending,sarkar2020exploring}, using MAP-Elites could help produce a wide variety of blended game levels as well as identify if certain regions of the blended latent space correspond to specific combinations of games being blended.

\section{Method}

\subsection{Level Data}
We tested our approach using levels of 5 classic NES-era platformers---\textit{Super Mario Bros. (SMB)}, \textit{Kid Icarus (KI)}, \textit{Mega Man (MM)}, \textit{Castlevania (CV)} and \textit{Ninja Gaiden (NG)}---taken from the Video Game Level Corpus (VGLC) \cite{summerville2016vglc}. The VGLC uses a text-based level representation with each unique character mapping to a tile in the game. Levels are additionally annotated with the path of an A* agent tuned using the jump arcs of the corresponding games as determined in prior work studying the jump physics of various platformers \cite{summerville2017mechanics}. This helps in generating levels that are more playable \cite{summerville2016mariostring,snodgrass2017procedural,sarkar2020exploring}. To account for differences in dimensions and orientations across games, we used uniform 16x16 level segments from each game for training our models. For this, we padded SMB levels with 2 rows of background tiles and the horizontal portions of MM levels, CV levels and NG levels with 1, 5 and 5 such rows respectively, obtaining 2643 segments for SMB, 1142 for KI, 2983 for MM, 3961 for CV and 3350 for NG. For training on the blended domain of SMB-KI-MM, we upsampled the SMB and KI segments to be the same as the number of MM segments to prevent the model from skewing towards MM. As we describe later, playability in the blended domain is determined by running an agent for each of the games in that blend, on each segment. Thus, we opted to use a 3-game blend combining SMB, KI and MM rather than all 5 due to the added computational cost that would be incurred when running 5 playability tests on each segment for thousands of MAP-Elites generations.

\subsection{Variational Autoencoders (VAEs)}
We trained a VAE on each of the above games and the blended domain. VAEs \cite{kingma2013autoencoding} are latent variable generative models that learn continuous, latent representations of data. They consist of encoder and decoder neural networks that respectively learn to map from data to latent space and vice-versa. In being trained to generate levels from latent vectors, the decoder of the VAE effectively learns the genotype-to-phenotype mapping. While the latent space enables sampling and interpolation to generate levels, it can also act as a continuous search space for evolving content. This involves starting with a population of randomly sampled latent vectors and optimizing an objective to generate desirable levels. Prior works have performed such latent evolution using CMA-ES \cite{sarkar2019blending}. In our work, we apply MAP-Elites in this space.

\XTABLEqdcov

\subsection{MAP-Elites}
MAP-Elites \cite{mouret2015illuminating} is a QD algorithm that divides the search space into niches or cells based on a desired set of attributes referred to as behavior characteristics (BCs) and returns the locally optimal solution in each cell, as determined by a predefined fitness function. Dimensions correspond to characteristics that describe the behavior of individual solutions, independent of their objective fitness. Each cell thus corresponds to a different region of the behavior space. When applied to game levels, dimensions can capture different types of levels or different level properties with the objective ensuring that constraints such as playability are satisfied.

\subsubsection{Behavior Characteristics (BCs)}
For our experiments, we used three sets of behavior characteristics - two based on tile-based metrics that capture some property of a level segment and one based on the presence of game elements in a segment.
\begin{itemize}
    \item \textit{Density (DE) and Nonlinearity (NL):} We define \textit{Density} to be the number of tiles in a segment that aren't background or path tiles. We define \textit{Nonlinearity} as a measure of how well the segment's topology follows a straight line based on the linear regression error on fitting a line to the structures within it. Since each segment is of dimension 16x16, the maximum possible value for \textit{Density} is 256. For \textit{Nonlinearity}, we set the max possible value to be 64, setting it a little higher than the max such value seen in the training data for all games, to allow room for discovering new types of segments. Thus the ranges for \textit{Density} and \textit{Nonlinearity} were [0, 256] and [0, 64] respectively. This yielded an archive consisting of $257 \times 65=16{,}705$ cells.
    \item \textit{Symmetry (SYM) and Similarity (SIM):} \textit{Symmetry}, as the name suggests, is a measure of how symmetrical a segment is along both the horizontal and vertical axes. Horizontally, it is computed by looking at pairs of rows starting at the center and moving outward and summing up the number of row positions that have the same tiles. Similarly, this is computed using pairs of columns for vertical symmetry. The final \textit{Symmetry} value for a segment is the sum of the horizontal and vertical symmetry. Since there are 8 pairs of rows/columns, each consisting of 16 positions, the maximum possible value is $8 \times 16 + 8 \times 16 = 256$.
    \textit{Similarity} is a measure of how similar a segment is compared to segments in the training data (i.e. the original levels). We define this to be simply the sum of the number of rows and columns in the generated segment that appear in the training set. Thus, the maximum value for this is 32 since there are 16 rows and 16 columns. As a metric, it is similar to the plagiarism metrics as defined in \cite{snodgrass2020multi,summerville2018expanding}. Thus, the ranges for \textit{Symmetry} and \textit{Similarity} were [0, 256] and [0, 32] respectively, giving us an archive consisting of $257 \times 33 = 8{,}481$ cells.
    \item \textit{Game Elements (GE):} Since the type of elements differ significantly across different games, we used different archive formulations for each game. Thus, we focused on looking at whether MAP-Elites can discover segments containing different combinations of elements specific to that game, rather than compare archives across games as in the previous two BCs. We looked at the following types of elements for each game:
    \begin{itemize}
        \item \textit{SMB:} Enemies, Pipes, ?-Marks, Collectables, Breakables
        \item \textit{KI:} Hazards, Doors, Moving Platforms, Fixed Platforms
        \item \textit{MM:} Hazards/Enemies, Doors, Ladders, Platforms, Collectables
        \item \textit{CV:} Hazards/Enemies, Doors, Ladders, Weapons, Collectables, Moving Platform, Breakable Wall
        \item \textit{NG:} Human Enemy, Animal, Ladder, Weapons, Collectables
        \item \textit{SMB-KI-MM Blend:} Enemy/Hazard, Door, Ladder, SMB Pipes, SMB ?-Marks, Collectables, Moving Platforms, Fixed Platforms, Breakables
    \end{itemize}
    Under this BC, each archive cell was represented as an N-digit binary number with 0/1 indicating the absence or presence of the corresponding element, with N being the number of element types considered for that game. For e.g. cell 11000 for SMB defines the space of segments containing enemies and pipes but no ?-marks, collectables or breakables. For each game/domain, the element archive thus had $2^N$ cells. Hence, we had $2^5=32$ cells for SMB, MM and NG, $2^4=16$ cells for KI, $2^7=128$ cells for CV and $2^9=512$ cells for the blend.
\end{itemize}

\subsubsection{Fitness}
For our fitness function, we used playability as determined by game-specific A* agents tuned using the jump arcs for the respective games, derived in prior work by Summerville et al. \cite{summerville2017mechanics} and used in \cite{sarkar2020exploring}. The fitness value is how far in the segment the agent can progress, normalized between 0 and 1. Thus, 1 indicates that the agent was able to fully traverse the segment using the jump arcs for that particular game. For SMB and CV, progress was determined only along the horizontal direction i.e. the x-axis, for KI only along the vertical direction i.e. y-axis, while for MM and NG, we considered both horizontal and vertical directions i.e. both axes. For the blended SMB-KI-MM domain i.e. \textit{Blend-Elites}, we tested playability of a segment by running each of the 3 agents and setting the fitness to be the highest among the values returned by the agent.

\subsection{VAE-MAP Elites}
Our algorithm for combining VAEs with MAP-Elites starts with an initial population of vectors sampled randomly from the VAE latent space and then assigning each to a cell based on the BCs as defined previously. Then, we run evolution for the desired number of generations. For each generation, we select parents by choosing 2 map cells at random and perform uniform crossover and mutation to produce the child. We then forward the resulting latent vector through the VAE decoder to obtain the level segment and determine the appropriate cell for the child based on the chosen BCs. That is, we either check for the presence of certain elements in the decoded segment or compute the appropriate metrics and assign a cell accordingly. We also run the agent-based playability objective on the segment to compute the fitness score. If the assigned cell is empty, we add the child vector and score. Else, we do so only if the score is greater than that of the existing vector in that cell. This procedure is then repeated for the remaining number of generations.

\section{Experiments}

All VAE models were implemented in PyTorch \cite{paszke2017automatic} and consisted of encoders and decoders of 4 fully-connected layers each, with ReLU activation. Models were trained for 10000 epochs with the Adam optimizer and a learning rate of 0.001 decayed every 2500 epochs by 0.01. We used a latent dimension size of 32 for each model. Model architecture was determined based on prior VAE-based PCGML works that utilized similar architectures. Initial experiments with bigger and smaller latent sizes did not have any noticeable impact on results, so we settled on a latent size of 32, similar to \cite{fontaine2020illuminating}.

For each game, we ran a trial for each of the 3 BCs, resulting in a total of 18 separate experiments. For each experiment, MAP-Elites was run for 100,000 generations, using a mutation probability of 0.3. For evaluation, we looked at the QD-score, Coverage and Optimality percentage. QD-score \cite{pugh2015confronting} is a standard evaluation metric for QD algorithms which sums the fitness values of all occupied cells in the archive. Coverage represents the percentage of cells in the archive that were occupied at the end of the run, thus indicating how much of the search space MAP-Elites was able to find a solution for during the run. Optimality further looks at what percentage of cells for which a solution was found had an optimal fitness value which in our case is playability. Additionally, for \review{\textit{Blend-Elites}}, we wanted to see if there are regions of the search space where certain agents and/or combinations of agents do better. If levels from certain regions are playable by a certain set of agents, that may indicate that those regions blend those specific games. Thus, for each cell in the archive, we kept track of which agents were able to complete a segment that was assigned to that cell.

\XFIGUREdenlplots

\XFIGUREsymsimplots

\XFIGUREelemplots

\XFIGUREdenl

\XFIGUREsymsim

\XFIGUREblendplots

\section{Results and Discussion}
Results for all experiments are given in Table \ref{XTABLEqdcov}. Progressions of QD-score and Coverage values over the course of the runs are shown in Figures \ref{XFIGUREdenlplots}--\ref{XFIGUREelemplots}. For both tile-based BCs, the highest QD-Score and Coverage values were observed for the blended SMB-KI-MM domain \review{i.e. \textit{Blend-Elites}}, followed closely by MM. In both cases, lowest values were obtained for CV and NG with KI and SMB doing better. Additionally, in most cases, if a solution was found for a cell, then the optimal solution was also found as exhibited by the high values for optimality percentage in all cases except \textit{Density-Nonlinearity} for KI. This is visualized in Figures \ref{XFIGUREdenl} and \ref{XFIGUREsymsim}.

\subsubsection{Tile-based BCs}
For \textit{Density-Nonlinearity}, MAP-Elites covers a smaller region in the left corner for SMB, CV and NG compared to the much larger coverage for KI and especially MM and \review{\textit{Blend-Elites}}. Levels in SMB, CV and NG tend to be less dense and more open than those in MM and KI and this bears out in these results. MAP-Elites finding the most number of playable levels in the archive for MM can be explained by the fact that MM has the most capable agent in terms of being able to move horizontally as well as in both directions vertically, thus enabling it to play through levels of more varied topologies. The KI agent also has vertical movement but only upwards and cannot move horizontally. While the NG agent also exhibits both horizontal and vertical movement, it moves only upward. Additionally, the horizontal sections of NG levels far outnumber the vertical ones in the dataset, and thus the model generates primarily horizontal segments, causing its behavior to be more aligned with the horizontal-only games like SMB and CV. Unsurprisingly, the orientation and movement models for the platformers have a significant impact on the range and diversity of playable levels that can be generated for that game. Moreover, the archive for \review{\textit{Blend-Elites}} seems to roughly intersect the regions covered by the three games (SMB, KI and MM) individually. For e.g., the lower left-corner region that is mostly unplayable by KI, is playable \review{for \textit{Blend-Elites}} likely because it is playable in SMB and MM. Much of these observations hold for \textit{Symmetry-Similarity} as well. MAP-Elites does best for MM and KI in terms of covering the archive with KI doing slightly better in finding playable levels dissimilar to training levels and MM doing a lot better in finding playable symmetric levels.

\subsubsection{Game Elements BC}
Since for this we used different archive sizes and definitions specific to individual games, we found it unsuitable to compare archives as in other BCs. Instead, we note that MAP-Elites was able to find elites for a high percentage of cells for all games, as indicated by the high Coverage values for this BC in Table \ref{XTABLEqdcov}, reaching 100\% for SMB, KI and NG. Additionally, in all cases, optimally playable solutions were found. This suggests an alternative application of MAP-Elites. A designer wishing to obtain a level that contains specific elements of a game could simply pick out such a level from the archive since it contains a playable level (if it exists) for each combination of elements for that game.

\subsubsection{\review{Blend-Elites}}
We further wanted to analyze the archive for the blended domain in order to see if different regions corresponded to different combinations of games being blended. One way to estimate this is by checking which agents can solve a segment from a given region of the archive. For example, if only the SMB agent is able to complete a segment but the KI and MM agents fail, then that segment is likely an SMB-like segment. However, if a segment is completed by both KI and MM agents but not SMB, then it is reasonable to expect that the segment blends both KI and MM but not SMB. Thus, for each cell in the archive, we kept track of the agents that managed to complete a segment that was assigned to that cell over the course of the entire MAP-Elites run. These results for both of the tile-based BCs are visualized in Figure \ref{XFIGUREblendplots}. For both BCs, we see distinct regions of blending emerge in terms of agent-based playability. For \textit{Density-Nonlinearity}, the lower left corner is playable only by SMB and MM but not KI, suggesting that region to be an SMB-MM blend. Then, as one moves outward towards the top-right corner, we encounter a region where levels are playable by all 3 agents suggesting this to be a region that blends all 3 games, followed by a region that mostly blends KI and MM and followed by an outer MM-only region. Similar observations can also be made with respect to the \textit{Symmetry-Similarity} plot. Combining MAP-Elites with VAEs thus offers some interesting implications for game blending moving forward. In addition to producing a diverse range of playable blended levels in one run, performing this agent-based analysis of the archive helps us identify different types of blend regions. In the future, this can allow designers to make levels based on explicit blend preferences. For e.g., a designer may want to build a level that initially blends SMB and KI, followed by a MM-only section and finish with segments blending all three games. In identifying the archive regions that contain specifically these types of segments, such applications can be made possible in the future.

\subsubsection{Examples}
Example levels for all 3 BCs for each game are shown in Figures \ref{XFIGUREdenllevels}-\ref{XFIGUREelemlevels}. Each example is an elite from the respective archive. For Figures \ref{XFIGUREdenllevels} and \ref{XFIGUREsymsimlevels}, each row roughly depicts 5 levels in the order of 1) low values for both BC dimensions, 2) high for the first/low for the second, 3) medium for both, 4) low for the first/high for the second and 5) high for both. Note that both metrics are computed based on the tile content of the segment, and so the values are sometimes related to each other, making it hard to precisely follow the above order and requiring some amount of cherry-picking. For game elements, we handpicked 5 labels for each game based on our preferences. For all BCs, examples for \review{\textit{Blend-Elites}} were picked from those cells whose elites were playable by all 3 agents.

\vspace{-0.1cm}
\section{Conclusion and Future Work}
We presented a PCGML approach that combines VAEs with the MAP-Elites algorithm for generating and blending game levels. This enabled the generation of a diverse range of playable levels in addition to identifying levels that blend specific combinations of games. There are several directions to consider in the future. 

In this work, we focused on comparing the application of MAP-Elites in several different games, rather than compare different QD algorithms. Thus, we would like to study other QD algorithms such as Novelty Search with Local Competition \cite{lehman2011evolving}, Constrained Novelty Search \cite{liapis2015constrained} or Constrained Surprise Search \cite{gravina2016surprise} when combined with VAEs. We could also try variations of the approach in this work, such as storing multiple elites per cell or combining MAP-Elites with more advanced models such as conditional VAEs and GMVAEs. A limitation of our approach is that we work with level segments rather than whole levels, a common practice in PCGML to overcome data scarcity. While this is less of a concern for games like Mario where segments can be automatically stitched together to yield entire levels, this isn't always feasible for games progressing in multiple directions and especially in blended domains. Prior work \cite{sarkar2020sequential} has addressed this issue by utilizing a VAE-based sequential model which in the future could be combined with our present approach. Note that even without the ability to generate whole levels, a designer could still choose from the archive of segments to manually construct a whole level as per their preference. In the future, it would also be worthwhile to conduct user studies and playtests to see if players perceive the generated levels to be sufficiently diverse. Finally, we would also like to incorporate MAP-Elites into ML-based co-creative and automated game design tools such as that presented in \cite{sarkar2020towards}.

\XFIGUREdenllevels

\XFIGUREsymsimlevels

\XFIGUREelemlevels


\bibliographystyle{ACM-Reference-Format}
\bibliography{refs-custom}

\end{document}